\documentclass[preprint,review,12pt]{elsarticle}
\usepackage{amssymb, amsmath, amsthm, bm}
\usepackage[noend]{algpseudocode}
\usepackage{algorithmicx,algorithm}
\usepackage{graphicx}
\usepackage{caption}
\usepackage{epstopdf}
\usepackage[colorlinks,linkcolor=blue]{hyperref}
\usepackage{bbding}
\usepackage{booktabs}
\usepackage{tikz,xcolor}

\usepackage{framed,multirow}
\usepackage{latexsym}
\usepackage{subfigure}
\usepackage{caption}
\usepackage{url}

\journal{}

\begin{document}
\begin{frontmatter}
\title{Physics-Informed Generator-Encoder Adversarial Networks with Latent Space Matching for Stochastic Differential Equations}
\author[label1] {Ruisong Gao \fnref{cor1}}
\author[label1] {Min Yang \corref{cor2}}
\author[label2] {Jin Zhang\fnref{cor3}}
\fntext[cor1] {Email: grs1108@163.com}
\cortext[cor2] {Corresponding author: yang@ytu.edu.cn}
\fntext[cor3] {Email: jinzhangalex@hotmail.com}

\address[label1]{School of Mathematics and Information Sciences, Yantai University, Yantai, China}
\address[label2]{Department of Mathematics, Shandong Normal University, Jinan, China}
		
\begin{abstract}
We propose a new class of physics-informed neural networks, called Physics-Informed Generator-Encoder Adversarial Networks,
to effectively address the challenges posed by forward, inverse, and mixed problems in stochastic differential equations.
In these scenarios, while the governing equations are known, the available data consist of only a limited set of snapshots for system parameters.
Our model consists of two key components: the generator and the encoder, both updated alternately by gradient descent.
In contrast to previous approaches of directly matching the approximated solutions with real snapshots,
we employ an indirect matching that operates within the lower-dimensional latent feature space.
This method circumvents challenges associated with high-dimensional inputs and complex data distributions,
while yielding more accurate solutions compared to existing neural network solvers.
In addition, the approach also mitigates the training instability issues encountered in previous adversarial frameworks in an efficient manner.
Numerical results provide compelling evidence of the effectiveness of the proposed method in solving different types of stochastic differential equations.
\end{abstract}

	
\end{frontmatter}

\section{Introduction}
Stochastic differential equations (SDEs) belong to a category of differential equations used to model the evolution of stochastic processes.
They combine theories of stochastic processes and differential equations to simulate phenomena imbued with randomness,
such as financial market fluctuations and noise in physical systems.
These equations find extensive applications in various domains, including finance, physics, biology, engineering, and more.

Numerical methods are typically employed to tackle the challenges of solving stochastic differential equations.
Techniques like Euler's method, Milstein's method, and the random Runge-Kutta method are commonly employed.
These methods discretize time steps and incorporate random numbers to simulate the unpredictable components.
For instance, Higham et al. \cite{Higham2005} introduced and analyzed two implicit methods for solving Ito stochastic differential equations featuring Poisson-driven jumps.
In their study, the first method, denoted as SSBE, is an expansion of the backward Euler strategy.
The second approach, termed CSSBE, leverages a compensated martingale representation of the Poisson process.
In the realm of finance and stochastic control, Yang et al. \cite{Yang2015} contributed a convergence analysis for specific types of forward-backward stochastic differential equations (FB-SDEs).
Sun et al. \cite{Sun2018} introduced pioneering second-order numerical schemes for first-order and second-order forward-backward stochastic differential equations (FBSDEs) using the Feynman-Kac formula.
Zhu et al. \cite{Zhu2019} delved into a weak Galerkin finite element method employing Raviart-Thomas elements, suited for a particular class of linear stochastic parabolic partial differential equations with additive noise across space and time.
Their work culminated in optimal strong convergence error estimates in the $L^2$ norm.
However, traditional numerical approaches may falter in addressing high-dimensional stochastic partial equations and can succumb to the challenges of the curse of dimensionality.

In recent years, remarkable progress has been made in solving differential equations through the integration of deep learning techniques \cite{Han2018,Liu2023,Raissi2018,Schaeffer2017,Sirignano2018,Tang2023,Wilson2022,Yang2022,Zhang2023}.
Particularly noteworthy is the emergence of the physics-informed neural network (PINN) paradigm \cite{Feng2022,Jagtap2020,Jagtap2022,Karumuri2021,Liu2021,Raissi2019,Yang2019},
which seamlessly integrating physical insights as adaptable constraints within the loss function.
This innovative approach leverages machine learning techniques such as automatic differentiation \cite{Baydin2017} and gradient descent for effective training.
In a previous work, Wang and Yao \cite{Wang2021} proposed a VNPs-SDE model, effectively handling noisy irregularly sampled data and providing robust uncertainty estimates.
This model fuses a modified form of neural processes to manage the noisy irregular data and incorporates SDE-Net for full data processing.
Furthermore, Wang et al. \cite{Wang2022} introduced a two-stage neural network-based approach for parameter estimation in parameterized stochastic differential equations driven by L\'{e}vy noise.
The application of deep learning also extends to the realm of quasi-linear parabolic stochastic partial differential equations,
where Teng et al. \cite{Teng2022} devised a deep learning-based numerical algorithm to solve forward-backward doubly stochastic differential equations,
including high-dimensional cases.
In scenarios where exact analytical parameter representations are lacking, and data from sparse sensors is limited,
Zhong et al. \cite{Zhong2023} employed variational autoencoders for solving the forward, inverse, and mixed problems of stochastic differential equations.
In addition, Guo et al. \cite{Guo2022} presented a data-driven approach for stochastic differential equations, utilizing a normalized field flow-based method.
A significant contribution came from Liu et al. with the introduction of PI-WGAN \cite{Liu2020},
which effectively addressed challenges in solving SDE problems by combining generative adversarial networks (GANs) \cite{Goodfellow2014} with physical insights.
This represents a pioneering use of adversarial training in the SDE domain, but faces challenges in terms of training stability.
To address such deficiencies, Gao et al. \cite{Gao2023Wang} introduced PI-VEGAN, which integrates an encoder to better capture the real data distribution.
However, the improvements in stability and accuracy come at the expense of longer training times and increased computational costs.

In this paper, we present a new class of physics-informed neural networks,
called physics-informed generator-encoder adversarial (PI-GEA) networks, for the effective and accurate solution of stochastic differential equations.
It is noted that current deep learning methods for solving stochastic equations primarily focus on minimizing the discrepancy between approximated solutions and collected snapshots.
While the concept of fitting generated approximations to real snapshots is intuitive,
it faces challenges, particularly when dealing with high-dimensional data characterised by complex distributions.
On the one hand, inaccuracies can arise due to the complexity nature of the data distribution, making it difficult to accurately capture the underlying patterns hidden in the snapshots.
On the other hand, training instability may occur due to the intricate interaction between high dimensionality and complex data distributions,
hampering the optimisation process of the model.
As a consequence, these challenges can collectively undermine the quality and reliability of the approximated solutions.
Taking inspiration from \cite{Ulyanov2018}, we develop an alternative method that uses an indirect matching strategy in the latent space,
with the primary goal of improving the accuracy and stability of physics-informed neural networks in solving stochastic differential equations.

Our model incorporates physical laws into a dual-network architecture that consists of two basic components: the generator and the encoder.
The generator module takes concatenated random noise and spatial coordinates as input and outputs approximated snapshots.
Simultaneously, the encoder module is designed to process both real and approximated snapshots, generating corresponding latent features.
These latent features play a key role in our methodology.
We proceed to evaluate the proximity of these latent features to a standard Gaussian distribution using the maximum mean discrepancy measurement.
This assessment provides insights into the distributional properties of the latent space and guides the optimization process.
As the latent features lie in a more informative lower-dimensional space compared to the original input data,
the proposed method not only addresses challenges associated with high dimensionality and intricate data distributions
but also improves the accuracy of solving stochastic equations compared to existing approaches.
Experimental results validate the satisfactory accuracy improvement brought by PI-GEA in solving forward, inverse, and mixed SDE problems.
Moreover, the proposed approach effectively addresses the training instability issues encountered in previous adversarial frameworks.

The remaining sections of this paper are structured as follows.
Section 2 outlines the problem types to be tackled and introduces the fundamental concept of adversarial training.
In Section 3, we delve into the details of the proposed approach for modeling stochastic processes and efficiently solving stochastic differential equations.
Section 4 is devoted to presenting the outcomes of our numerical experiments.
Concluding the paper, Section 5 discusses the limitations of our method and outlines potential avenues for future research.

\section{Background}
\subsection{Problem Setup}
Let $\Omega$ be a probability space, and $\omega$ be a random event. We consider the following stochastic differential equation:
\begin{align}
	\label{eq1}
	\begin{split}
		\mathcal{N}_x[u(x; \omega),k(x; \omega)] & = f(x; \omega),
		\quad x \in \mathcal{D},\quad\omega \in \Omega,
		\\[5pt]
		\mathcal{B}_x[u(x; \omega)] & = b(x; \omega),
		\quad x \in \Gamma,
	\end{split}
\end{align}
where $x$ represents the spatial coordinate, and $\mathcal{D}$ is a physical domain in $\mathbb{R}^d$.
The operator $\mathcal{N}_x$ denotes a general differential operator,
while $\mathcal{B}_x$ represents the operator acting on the domain boundary $\Gamma$.
The coefficients $k(x; \omega)$, forcing term $f(x; \omega)$, and boundary condition $b(x; \omega)$ are all stochastic processes.
Consequently, the solution $u(x; \omega)$ is also a stochastic process, depending on $k(x; \omega)$ and $f(x; \omega)$.

To gather measurements of the stochastic process described in \eqref{eq1},
it is common to deploy sensors uniformly throughout the domain.
The coordinates of these sensors for $u(x; \omega)$, $k(x; \omega)$, $f(x; \omega)$, and $b(x; \omega)$ are denoted as $\{x_i^u\}_{i=1}^{n_u}$, $\{x_i^k\}_{i=1}^{n_k}$, $\{x_i^f\}_{i=1}^{n_f}$,
and $\{x_i^b\}_{i=1}^{n_b}$, respectively, where $n_u$, $n_k$, $n_f$, and $n_b$ represent the corresponding number of sensors.
It is assumed that there are sufficient sensor measurements available for $f(x; \omega)$, given the known boundary condition $b(x; \omega)$.
Depending on the available measurements of $k(x; \omega)$ and $u(x; \omega)$, three distinct types of problems can be identified: the forward problem, the inverse problem, and the mixed problem.

The forward problem aims to approximate the solution $u(x; \omega)$ given the measurements of $k(x; \omega)$.

The inverse problem involves estimating the coefficient $k(x; \omega)$ given the measurements of $u(x; \omega)$.

The mixed problem requires computing the solution $u(x; \omega)$ and the coefficient $k(x; \omega)$ simultaneously,
with only partial knowledge of $u(x; \omega)$ and $k(x; \omega)$ available.

As the number of sensors measuring $u(x; \omega)$ increases while the number of sensors measuring $k(x; \omega)$ decreases,
the nature of the estimation gradually transitions from a forward problem to a mixed problem, and eventually to an inverse problem.

Given a random event $\omega \in \Omega$, we employ distributed sensors to capture snapshots of the stochastic processes.
Assuming we possess a set of $N$ snapshots represented by:
\begin{align}
	\label{RealSnap}
	\{H(\omega^{(j)})\}_{j=1}^N = \{(K(\omega^{(j)}), U(\omega^{(j)}), F(\omega^{(j)}), B(\omega^{(j)}))\}_{j=1}^N,
\end{align}
where
\begin{align*}
	K(\omega^{(j)}) &= (k(x_i^k; \omega^{(j)}))_{i=1}^{n_k}, \quad
	U(\omega^{(j)}) = (u(x_i^u; \omega^{(j)}))_{i=1}^{n_u},
	\\[5pt]
	F(\omega^{(j)}) &= (f(x_i^f; \omega^{(j)}))_{i=1}^{n_f},\quad
	B(\omega^{(j)}) = (b(x_i^b; \omega^{(j)}))_{i=1}^{n_b}.
\end{align*}
In the equation \eqref{RealSnap}, the relevant terms are omitted when there are no sensors for a specific process.
For the forward problem, we set $n_u = 0$, signifying the absence of sensors for the solution $u(x; \omega)$.
For the inverse problem, we set $n_k = 1$, indicating the presence of a single sensor measuring the coefficient $k(x; \omega)$.

Our approach falls within the realm of adversarial neural networks, a category of machine learning methodologies
where two or more sets of neural networks engage in both mutual competition and cooperation throughout the training process.
These networks progressively enhance their performance through adversarial interactions.
Next, we provide a brief overview of the concept of adversarial training.

\subsection{Adversarial Training in the Latent Feature Space}

One of the most prominent frameworks within adversarial neural networks is Generative Adversarial Networks (GANs).
GANs consist of two foundational networks: the generator $\mathcal{G}$ and the discriminator $\mathcal{D}$.
The generator takes  random noises as input and outputs synthetic data.
The main goal of the generator is to minimize the discriminator's ability to distinguish between real and synthetic data.
Conversely, the discriminator strives to maximize its accuracy in distinguishing between the actual and generated data.
This dynamic interplay between the two components during training gives rise to the term ``adversarial''.
Their optimization goals can be summarized as follows:
\begin{align}
	\min_{\mathcal{G}}\mathbb{E}_{\xi}[\log(1 - \mathcal{D}(\mathcal{G}(\xi)))],
\end{align}
\begin{align}
	\max_{\mathcal{D}}\mathbb{E}_{y}[\log \mathcal{D}(y)] + \mathbb{E}_{\xi}[\log(1 - \mathcal{D}(\mathcal{G}(\xi)))],
\end{align}
where $ y $ represents the real sample, and $\xi$  denotes random noise drawn from a specified prior distribution (e.g., standard Gaussian distribution).
Over successive iterations of this adversarial training process, the generator progressively improves its capacity to produce samples that closely resemble real data.

Recently, Liu et al. \cite{Liu2020} introduced PI-WGAN,
a method that combines generative adversarial networks with physical insights to address forward, inverse, and mixed stochastic differential equation problems.
Despite showcasing remarkable approximation capabilities, PI-WGAN faces the challenge of training instability.

In order to mitigate the instability within the PI-WGAN framework,
our previous research \cite{Gao2023Wang} introduced an additional variational encoder for approximating the latent variables of the actual distribution of the measurements.
These latent variables are integrated into the generator to facilitate accurate learning of the characteristics of the stochastic partial equations.
While this enhanced approach demonstrates notable improvements in terms of training stability and accuracy, it does indeed come with certain trade-offs.
One trade-off is the increased complexity of the network architecture.
The introduction of an additional variational encoder and its integration with the generator can lead to a more complicated model structure.
This complexity  can have an impact on factors such as computational resources, memory requirements, and overall model manageability.
Another trade-off is the increase in training time.
The incorporation of the variational encoder and the subsequent adjustments to the learning process of the generator can result in longer training durations.

Note that in a generative adversarial network framework,
the discriminator plays a role in measuring the distance between the actual distribution and the generated distribution.
This distance measurement occurs directly in the high-dimensional space of the training data.
In \cite{Ulyanov2018},
the authors utilized an encoder to substitute the traditional discriminator. This encoder fulfills the role of transforming data into lower-dimensional vectors.
This transformation enables an indirect comparison, or matching, within the latent feature space.

To be precise, let $\mathcal{E}$ represent the encoder module, designed in a manner that its output dimension is notably smaller than the input dimension.
The primary role of the encoder is to convert high-dimensional and complex data into a lower-dimensional representation that adheres to a simpler distribution.
Specifically, the encoder $\mathcal{E}$ receives either an actual sample $y$ or a synthetic sample $\mathcal{G}(\xi)$ from the generator,
and then produces the low-dimensional feature representation for the input, denoted as $z_{real}$ and $z_{gen}$, respectively.

Denote by $\textrm{MMD}(\cdot,\cdot)$ the Maximum Mean Discrepancy (MMD) loss,
which measures the proximity between two probability distributions and is defined as \cite{Gretton2012}:
\begin{align*}
\text{MMD}(P, Q) = \frac{1}{n^2} \sum_{i=1}^{n} \sum_{j=1}^{n} k(x_i, x_j) - \frac{2}{nm} \sum_{i=1}^{n} \sum_{j=1}^{m} k(x_i, y_j) + \frac{1}{m^2} \sum_{i=1}^{m} \sum_{j=1}^{m} k(y_i, y_j)
\end{align*}
where $k(\cdot, \cdot)$ is the Gaussian kernel function,
and $\{x_i\}_{i=1}^n $ and $\{y_j\}_{j=1}^m $ refer to samples from the distributions $P$ and $Q$, respectively.

The interplay between the generator module and the encoder module is now determined by the optimization objectives outlined as follows:
\begin{align}
\label{Gloss}
	\min_{\mathcal{G}}\underset{\text{Adversarial loss}}{\underline{\mathbb{E}_{z_{gen}, z}[\textrm{MMD}(z_{gen}, z)]}} +
	\underset{\text{Generation loss}}{\underline{\mathbb{E}_{\xi, y}[\textrm{MMD}(\mathcal{G}(\xi), y)]}},
\end{align}
\begin{align}
\label{Eloss}
	\max_{\mathcal{E}}\underset{\text{Adversarial loss}}{\underline{\mathbb{E}_{z_{gen}, z_{real}, z}[\textrm{MMD}(z_{gen}, z) - \textrm{MMD}(z_{real}, z)]}} -
	\underset{\text{Reconstruction loss}}{\underline{\mathbb{E}_y[\textrm{MMD}(\mathcal{G}(\mathcal{E}(y)), y)]}},
\end{align}
for any $ z \sim N(0, I) $.

From the above optimization objectives, we can see that the generator's primary task is to produce synthetic data that resembles the distribution of real data.
The adversarial loss component seeks to minimize the mean discrepancy between the low-dimensional feature $z_{gen}$ of synthetic data and a random noise sample $z$,
which encourages the synthetic data distribution to closely align with the distribution of the random noise, enhancing the authenticity of the generated data.
The generation loss, on the other hand, emphasizes that the synthetic data should be concentrated in the low-dimensional feature space around the distribution of real data.

The encoder's main role is to map real data to a lower-dimensional feature space while minimizing the spread of low-dimensional features of synthetic data and real data.
The adversarial loss accentuates that the encoder should minimize the distributional difference between the low-dimensional features of synthetic data and those of random noise,
effectively reducing the dissimilarity between synthetic data and random noise in the low-dimensional space.
The reconstruction loss component evaluates the spread between the low-dimensional features of reconstructed data (after passing through the encoder and generator) and the original real data.
This underscores an additional role of the encoder to ensure that synthetic data can accurately be mapped back to the original data in the low-dimensional  latent feature space,
thereby enhancing the quality and coherence of the synthetic data.

By replacing the traditional discriminator with the encoder, the approach shifts the evaluation and comparison process from the high-dimensional data space to the more informative and condensed latent feature space. This introduces the benefits of indirect matching, ultimately contributing to enhanced data generation and synthesis within the model.

\section{Methodology}
In this section, we present two PI-GEA algorithms:
one aimed to approximate stochastic processes, and another designed for solving stochastic differential equations.
While the primary emphasis of PI-GEA is not on the approximation of stochastic processes,
numerical experiments to reconstruct the sample paths of such processes allow a preliminary assessment of its performance in a stochastic environment.
We then delve into the construction and training procedures of PI-GEA, outlining its potential for efficiently solving forward, backward and mixed problems associated with SDEs.

\subsection{Approximation of stochastic process}
\label{Approximation_sp}
Suppose that we have a set of snapshots for the stochastic process $f(x; \omega)$ as follows:
\begin{align}
	\{F(\omega^{(j)})\}_{j=1}^N = \{(f(x_i; \omega^{(j)}))_{i=1}^{n_f}\}_{j=1}^N,
\end{align}
where $N$ represents the total number of snapshots,
$n_f$ is the count of scattered sensors,
and $\{x_i\}_{i=1}^{n_f}$ denotes the locations of these sensors.

We use a generator network $\tilde{f}_\theta $ parameterized by $ \theta $ to model the stochastic process $ f(x; \omega) $.
Let $\mathcal{E}_{\phi} $  be the encoder parameterized by $\phi$.

\textbf{Training process}.
In the training phase, the generator takes as input the concatenation of the spatial coordinate $x_i$
and an $m$-dimensional random noise $\xi_j$ sampled from a standard Gaussian distribution,
and then outputs the approximation $\tilde{f}_{\theta}(x_i;\xi_j)$  of $f(x_i; \omega^{(j)})$.
The set of generated snapshots at each training epoch is represented as follows:
\begin{align}
	\label{tilde_F}
	\{\tilde{F}(\omega^{(j)})\}_{j=1}^n = \{( \tilde{f}_{\theta}(x_i;\xi_j) )_{i=1}^{n_f}\}_{j=1}^n,
\end{align}
where $n$ represents the batch size.
On the other hand,
the encoder receives either $F(\omega^{(j)})$ or $\tilde{F}(\omega^{(j)})$ as its input and produces the corresponding latent feature representation,
denoted as $z_{real}^j$ and $z_{gen}^j$, respectively.
Let  $\{\hat{F}(\omega^{(j)})\}_{j=1}^n$ be the set of reconstructed snapshots defined as
\begin{align}
	\label{hat_F}
	\{\hat{F}(\omega^{(j)})\}_{j=1}^n = \{( \tilde{f}_{\theta}(x_i; z_{real}^j) )_{i=1}^{n_f}\}_{j=1}^n.
\end{align}

In view of \eqref{Gloss} and \eqref{Eloss},
the adversarial training objectives for modeling the stochastic process $ f(x; \omega) $ are formulated as:
\begin{align}
	\label{loss_G1}
	\min_{\theta}\frac{1}{n}\sum_{j=1}^{n}\bigg( \textrm{MMD}(z_{gen}^j, z^j) + \textrm{MMD}\big( \tilde{F}(\omega^{(j)}), F(\omega^{(j)}) \big) \bigg),
\end{align}
\begin{align}
	\label{loss_E1}
	\max_{\phi}\frac{1}{n}\sum_{j=1}^{n}\bigg(\textrm{MMD}(z_{gen}^j, z^j) - \textrm{MMD}(z_{real}^j, z^j) - \textrm{MMD}\big( \hat{F}(\omega^{(j)}), F(\omega^{(j)}) \big) \bigg),
\end{align}
where $z^j \sim N(0, I_m)$.
We iteratively update the network parameters using the gradient descent method,
and the algorithm details are presented in Algorithm \ref{Algorithm_1}.

\begin{algorithm}[hbt]
	\caption{Stochastic Process Approximation}
	\label{Algorithm_1}
	\begin{algorithmic}
		\State \textbf{Input}: Initial network parameters $\theta$ and $\phi$, number of iterations $epochs$, batch size $n$.
		\For{$ t = 1$ to $epochs$}
			\State Sample $N$ real snapshots $\{F(\omega^{(j)})\}_{j=1}^N$.
			\For{$ c = 1$ to $\frac{N}{n}$}
				\State Sample $n$ random noises $\{\xi_j\}_{j=1}^n$;
				\State Calculate real latent feature representations $\{z_{real}^j\}_{j=1}^n = \mathcal{E}_{\phi}(\{F(\omega^{(j)})\}_{j=1}^n)$;
				\State Compute reconstructed snapshots $\{\hat{F}(\omega^{(j)})\}_{j=1}^n$ using \eqref{hat_F};
				\State Generate synthetic snapshots $\{\tilde{F}(\omega^{(j)})\}_{j=1}^n$ using \eqref{tilde_F};
				\State Calculate latent feature representations $\{z_{gen}^j\}_{j=1}^n = \mathcal{E}_{\phi}(\{\tilde{F}(\omega^{(j)})\}_{j=1}^n)$;
				\State Update encoder parameters $\phi$ using \eqref{loss_E1};
				\State Generate $\{\tilde{F}(\omega^{(j)})\}_{j=1}^n$ with the updated encoder;
				\State Recalculate latent feature representations $\{z_{gen}^j\}_{j=1}^n = \mathcal{E}_{\phi}(\{\tilde{F}(\omega^{(j)})\}_{j=1}^n)$;
				\State Update generator parameters $\theta$ using \eqref{loss_G1}.
			\EndFor
		\EndFor
		\State \textbf{Output}: Network parameters $\phi$ and $\theta$.
	\end{algorithmic}
\end{algorithm}

\textbf{Test process}.
Assume that the test coordinates are denoted as $\{x_i\}_{i=1}^p$, where $p\gg n_f$.
Initially, $N$ random noises $\xi_j$ are sampled from a Gaussian distribution $N(0, I_m)$.
Subsequently, these samples are input into the trained generator $\tilde{f}_{\theta}$, resulting in the generation of $\{\tilde{F}(\omega^{(j)})\}_{j=1}^N$.
It is important to emphasize that $x_i$ within this context can represent any coordinate located within the physical domain $\mathcal{D}$.

\subsection{Solving SDEs with PI-GEA}

\begin{figure}[htb]
	\centering
	\scalebox{.8}{\includegraphics[width=0.9\linewidth]{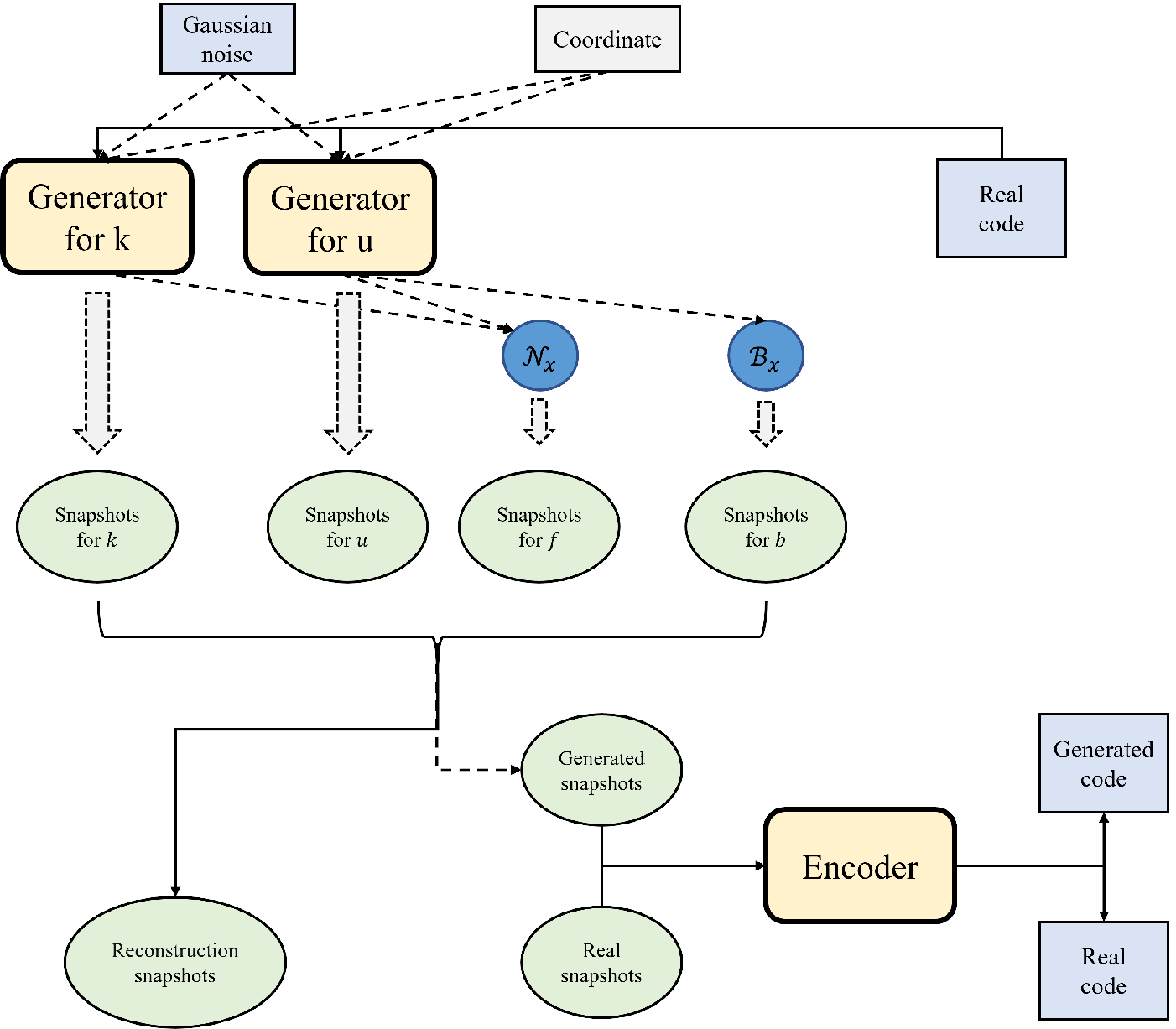}}
	\caption{The architecture of PI-GEA for solving stochastic differential equations.
		The brackets signify concatenation. During the training process, both the solid and dashed parts are engaged,
        whereas during the testing process, only the dashed portion is involved.}
	\label{PI-GEA}
\end{figure}

To solve a stochastic differential equation, such as \eqref{eq1},
we assume that we have the measurement data as described in \eqref{RealSnap}.

Within our network architecture, we incorporate two generators, $\tilde{k}_{\theta_k}$ and $\tilde{u}_{\theta_u}$,
alongside an encoder $\mathcal{E}_{\phi}$,
with $\theta_k$, $\theta_u$, and $\phi$ denoting the respective network parameters.
For a visual representation of the network architecture, please refer to Figure \ref{PI-GEA}.

\textbf{Training process}.
Initially, we approximate the stochastic processes $k(x; \omega)$ and $u(x; \omega)$ with the generators $\tilde{k}_{\theta_k}(x, \xi)$ and $\tilde{u}_{\theta_u}(x, \xi)$, respectively,
as described in Section \ref{Approximation_sp}.

Next, we integrate the physical equation \eqref{eq1} into the neural network architecture by applying the differential operators $\mathcal{N}_x$ and $\mathcal{B}_x$
to the generators $\tilde{k}_{\theta_k}(x, \xi)$ and $\tilde{u}_{\theta_u}(x, \xi)$, using automatic differentiation. This yields
\begin{align}
	\begin{split}
	\tilde{f}_{\theta_k, \theta_u}(x; \xi) &= \mathcal{N}_x [ \tilde{k}_{\theta_k}(x; \xi),  \tilde{u}_{\theta_u}(x; \xi)],\\
	\tilde{b}_{\theta_u}(x; \xi) &= \mathcal{B}_x [ \tilde{u}_{\theta_u}(x; \xi)].
	\end{split}
\end{align}
In this context, $\tilde{f}_{\theta_k, \theta_u}(x; \xi)$ and $\tilde{b}_{\theta_u}(x; \xi)$ represent neural networks
that serve to approximate the source term $f(x; \omega)$ and the boundary term $b(x; \omega)$, respectively.

Utilizing the generator networks, we can acquire a set of synthetic snapshots as follows:
\begin{align}
	\label{tilde_H}
	\{\tilde{H}(\omega^{(j)})\}_{j=1}^n = \{(\tilde{K}(\xi_j), \tilde{U}(\xi_j), \tilde{F}(\xi_j), \tilde{B}(\xi_j))\}_{j=1}^n,
\end{align}
where
\begin{align*}
	\tilde{K}(\xi_j) = (\tilde{k}_{\theta_k}(x_i^k; \xi_j))_{i=1}^{n_k}, \quad &\tilde{U}(\xi_j) = (\tilde{u}_{\theta_u}(x_i^u; \xi_j))_{i=1}^{n_u},\\
	\tilde{F}(\xi_j) = (\tilde{f}_{\theta_k, \theta_u}(x_i^f; \xi_j))_{i=1}^{n_f}, \quad  &\tilde{B}(\xi_j) = (\tilde{u}_{\theta_u}(x_i^b; \xi_j))_{i=1}^{n_b}.
\end{align*}

Furthermore, the encoder $\mathcal{E}_{\phi}$ accepts $H(\omega^{(j)})$ and $\tilde{H}(\omega^{(j)})$ as inputs
and subsequently generates their corresponding low-dimensional latent feature representations, denoted as $z_{real}^j$ and $z_{gen}^j$, respectively.

Let $\hat{H}(\omega^{(j)})$ be the set of reconstructed snapshot defined as
\begin{align}
	\label{hat_H}
	\{\hat{H}(\omega^{(j)})\}_{j=1}^n = \{(\tilde{K}(z_{real}^j), \tilde{U}(z_{real}^j), \tilde{F}(z_{real}^j), \tilde{B}(z_{real}^j))\}_{j=1}^n.
\end{align}
where
\begin{align*}
	\tilde{K}(z_{real}^j) = (\tilde{k}_{\theta_k}(x_i^k; z_{real}^j))_{i=1}^{n_k}, \quad &\tilde{U}(z_{real}^j) = (\tilde{u}_{\theta_u}(x_i^u; z_{real}^j))_{i=1}^{n_u},\\
	\tilde{F}(z_{real}^j) = (\tilde{f}_{\theta_k, \theta_u}(x_i^f; z_{real}^j))_{i=1}^{n_f}, \quad  &\tilde{B}(z_{real}^j) = (\tilde{u}_{\theta_u}(x_i^b; z_{real}^j))_{i=1}^{n_b}.
\end{align*}
The final optimization objectives in solving the stochastic equation \eqref{eq1} are formulated as:
\begin{align}
	\label{loss_G2}
	\min_{\theta_k, \theta_u}\frac{1}{n}\sum_{j=1}^{n}\bigg( \textrm{MMD}(z_{gen}^j, z^j) + \textrm{MMD}\big( \tilde{H}(\omega^{(j)}), H(\omega^{(j)}) \big) \bigg),
\end{align}
\begin{align}
	\label{loss_E2}
	\max_{\phi}\frac{1}{n}\sum_{j=1}^{n}\bigg(\textrm{MMD}(z_{gen}^j, z^j) - \textrm{MMD}(z_{real}^j, z^j) - \textrm{MMD}\big( \hat{H}(\omega^{(j)}), H(\omega^{(j)}) \big) \bigg),
\end{align}
where $z^j \sim N(0, I_m)$.

Note that the proposed framework can solve three types of problems:
the forward problem, the inverse problem, and the mixed problem.
and the algorithm details are shown in Algorithm \ref{Algorithm_2}.

\begin{algorithm}[hbt]
	\caption{Solving SDEs using PI-GEA}
	\label{Algorithm_2}
	\begin{algorithmic}
		\State \textbf{Input}: Initial network parameters $\theta_k$, $\theta_u$, and $\phi$, number of training iterations $epochs$, batch size $n$.
		\For{$ t = 1$ to $epochs$}
			\State Sample $N$ real snapshots $\{H(\omega^{(j)})\}^N_{j=1}$.
			\For{$ c = 1$ to $\frac{N}{n}$}
				\State Sample $n$ random noises $\{\xi_j\}_{j=1}^n$;
				\State Calculate real latent feature representations $\{z_{real}^j\}_{j=1}^n = \mathcal{E}_{\phi}(\{H(\omega^{(j)})\}_{j=1}^n)$;
				\State Compute reconstructed snapshots $\{\hat{H}(\omega^{(j)})\}_{j=1}^n$ using \eqref{hat_H};
				\State Generate synthetic snapshots$\{\tilde{H}(\omega^{(j)})\}_{j=1}^n$ using \eqref{tilde_H};
				\State Calculate latent feature representations $\{z_{gen}^j\}_{j=1}^n = \mathcal{E}_{\phi}(\{\tilde{H}(\omega^{(j)})\}_{j=1}^n)$;
				\State Update encoder parameters $\phi$ using \eqref{loss_E2};
				\State Generate $\{\tilde{H}(\omega^{(j)})\}_{j=1}^n$ with the updated encoder;
				\State Recalculate latent feature representations $\{z_{gen}^j\}_{j=1}^n = \mathcal{E}_{\phi}(\{\tilde{H}(\omega^{(j)})\}_{j=1}^n)$;
				\State Update generator parameters $\theta_k$ and $\theta_u$ using \eqref{loss_G2}.
			\EndFor
		\EndFor
		\State \textbf{Output}: Network parameters $\theta_k$, $\theta_u$, and $\phi$.
	\end{algorithmic}
\end{algorithm}

\textbf{Test Process}. Let us consider the scenario where the test coordinates are represented as $\{x_i\}_{i=1}^p$.
Subsequently, proceed by sampling $N$ random noises $\xi_j \sim N(0, I_m)$.
Following this, utilize the trained generators $\tilde{k}_{\theta_k}$ and $\tilde{u}_{\theta_u}$ to generate $\{\tilde{H}(\omega^{(j)})\}_{j=1}^N$.
It's important to emphasize that $x_i$ within this context has the potential to denote any coordinate positioned within the physical domain $\mathcal{D}$.

The framework we have presented is versatile in its ability to handle three types of stochastic equations: the forward problem, the inverse problem, and the mixed problem.
In contrast to previous deep learning solvers that measure the quality of generated approximations against real snapshots,
the proposed approach shifts the evaluation and comparison process from the high-dimensional data space to a more informative and condensed latent feature space.
This strategic introduces the benefits of indirect matching, which, in turn, enhances the accuracy of approximations and contributes to the overall stability of the model during training.
Furthermore,  the absence of a discriminator leads to a more concise network architecture when compared to our previous work \cite{Gao2023Wang}.
As a result, it not only contributes to solution accuracy but also mitigates the computational load, resulting in a more efficient training process.

\section{Numerical results}
\subsection{Experimental Setup}
In this section, we evaluate the performance of our model through a series of numerical experiments,
which are divided into two primary components: approximating stochastic processes and solving stochastic differential equations.
We conduct a thorough comparative analysis between our approach and several baseline methods,
specifically PI-VAE \cite{Zhong2023}, PI-WGAN \cite{Liu2020}, and PI-VEGAN \cite{Gao2023Wang}.
All methods are implemented using the PyTorch framework on an Intel CPU i7-8700 platform with 16 GB of memory and a single GTX 1660 GPU.

To ensure the smoothness of solutions for higher-order derivatives,
we adopt the Tanh function as the activation function for the network.
Unless explicitly stated, we set the dimension of the random noise (output dimension of the encoder) to 4, the learning rate to 0.0001, the batch size $n$ to 500, the number of training processes to 10000,
and configure both the generator and the encoder with four hidden layers, each having a width of 128, as the default settings.

We simulate sensor measurements using a Monte Carlo method and employ a finite difference scheme to generate both the training and test data.
Additionally, we reference 1000 sample paths for resampling.
For our experiments, we set the number of test points $p$ to 101 and generate 1000 test samples.
We evaluate the model's performance using the following metrics.

\textbf{Wasserstein Distance} \cite{Cuturi2013}:
This metric quantifies the dissimilarity between the real distribution (collected snapshots) and the generated distribution (samples produced by the generator).
A smaller Wasserstein distance indicates a closer match between the two distributions.
The formula for the Wasserstein distance is defined as:
\begin{align*}
W(P, Q) = \inf_{\gamma\sim\prod(P,Q)}\mathbb{E}_{x,y\sim\gamma}[|x-y|]
\end{align*}
Here, $\prod(P,Q)$ represents the set of all possible joint distributions combining the distributions $P$ and $Q$.

\textbf{Principal Component Analysis (PCA) }\cite{Bro2014}:
We assess the similarity of the underlying distributions by analyzing the eigenvalues of the covariance matrix.
PCA is applied to the observation matrix $X \in \mathbb{R}^{N \times M}$ to extract principal eigenvalues.
Similar eigenvalues between the generated and real distributions suggest a greater degree of similarity.

\textbf{Relative $L^2$ Error}: We uniformly select 30 models from the last 3000 epochs
to evaluate their average performance in terms of the relative $L^2$ error for the mean and standard deviation of the approximated stochastic process.
For a stochastic process $u(x;\omega)$, the mean and standard deviation can be represented as follows:
\begin{align*}
\mu(x) = \mathbb{E}_{\omega}[u(x; \omega)], \quad
\sigma(x) = \sqrt{\mathbb{E}_{\omega}[(u(x; \omega)-\mu(x))^2]}.
\end{align*}
Then the corresponding relative $L^2$ errors of the approximation $\hat{u}(x;\omega)$ are computed by
\begin{align}
	\frac{\|\hat{\mu}(x)-\mu(x)\|_2}{\|\mu(x)\|_2},
	\qquad
	\frac{\|\hat{\sigma}(x)-\sigma(x)\|_2}{\|\sigma(x)\|_2}.
	\label{relative_error}
\end{align}

\subsection{Performance Evaluation in Approximating Stochastic Processes}
Consider a Gaussian process with zero mean and squared exponential kernel:
\begin{align}
	f(x) \sim \mathcal{GP}(0, \exp(\dfrac{-(x-x')^2}{2l^2})),\quad x, x' \in [-1, 1],
\end{align}
Where $l$ represents the correlation length.

The number of sensors is set to 6.
For the correlation length, we consider three cases: $l=1$, $l=0.5$, and $l=0.2$.
We trained the model for 5000 epochs in each case.
Both the generator and encoder are configured with 3 hidden layers, each containing 64 neurons.

\begin{figure}[htb]
	\centering
		\includegraphics[height=4cm,width=16cm]{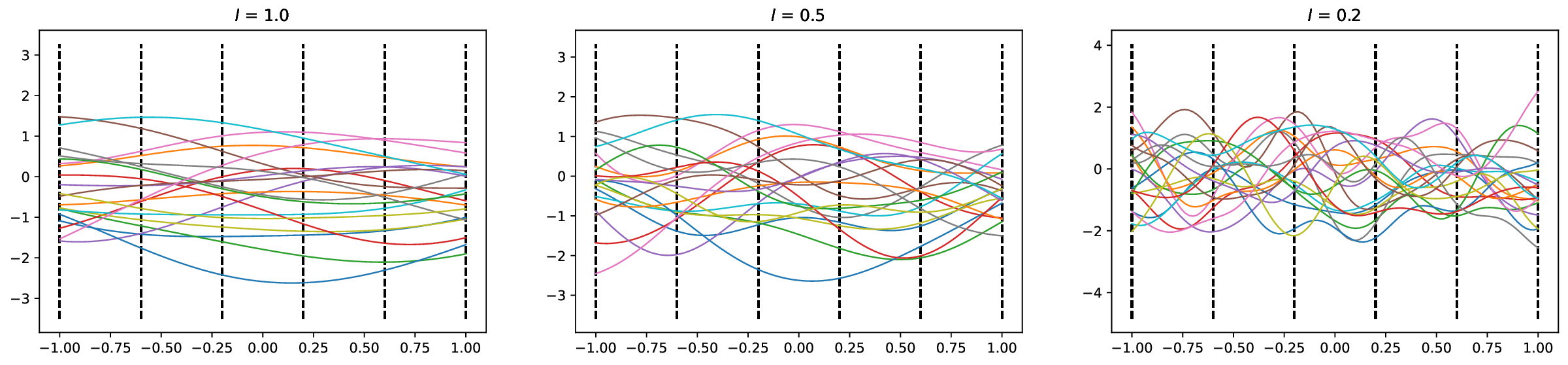}
	\caption{Training data for approximate Gaussian processes.
		The positions of the sensor are indicated by black vertical dotted lines.}
	\label{sample_sp}
\end{figure}

\begin{figure}[!htb]
	\centering
		\includegraphics[height=10cm,width=10cm]{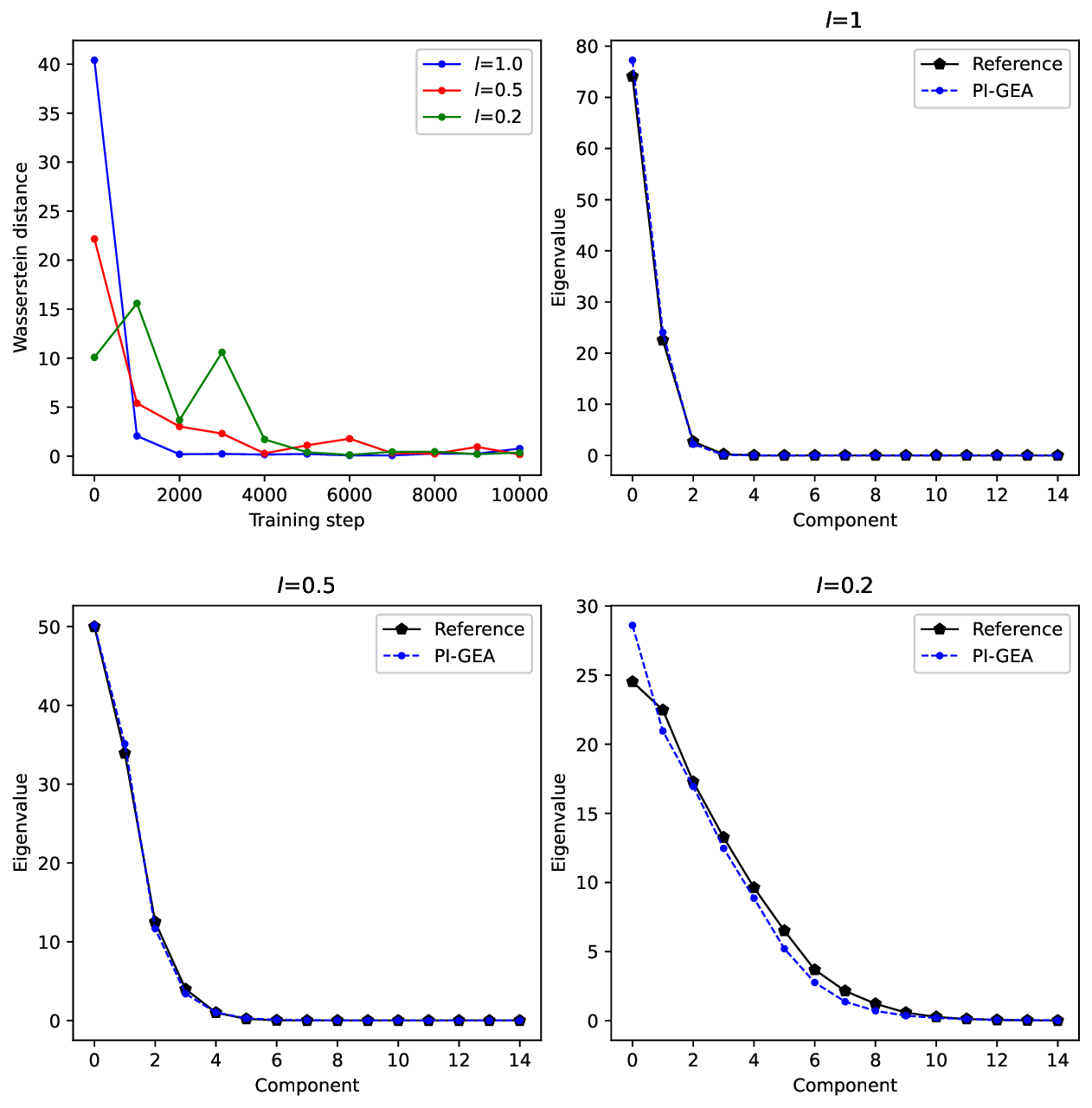}
	\caption{Approximating stochastic processes.
  The initial plot depicts the progression of the Wasserstein distance between the samples generated by our model and the reference samples throughout the training process.
  The subsequent three plots illustrate the eigenvalues of the sample covariance matrix.}
	\label{result_sp}
\end{figure}

As can be observed from the training data depicted in Figure \ref{sample_sp},
a smaller correlation length leads to more rapid changes or fluctuations in the training data,
which in turn poses greater difficulty for the approximation process.
Figure \ref{result_sp} illustrates the outcomes of our experiments in approximating stochastic processes.

We first evaluate the Wasserstein distance between the generated samples and the reference samples.
As depicted in Figure \ref{result_sp}, the Wasserstein distance converges and approaches zero as training progresses.
This trend indicates that the distribution of generated samples gradually aligns with the real distribution.

We proceed to compare the eigenvalues of the covariance matrix between the generated samples and the reference samples.
Remarkably, our model performs exceptionally well in both scenarios, where correlation lengths are $l=1.0$ and $l=0.5$, effectively capturing the local behavior of the target stochastic process.
Even for the more challenging scenario with $l=0.2$, our approach remains effective in approximating the stochastic process.

\subsection{Forward problem}
Starting from this section, we address the forward, inverse,
and mixed problems of stochastic differential equations.
Consider the following  elliptic stochastic differential equation:
\begin{align}
	\label{SDE_example}
	-\frac{1}{10}\frac{d}{dx}[k(x; \omega)\frac{d}{dx} u(x; \omega)] = f(x; \omega), \quad x \in [-1, 1],
\end{align}
where we impose homogeneous Dirichlet boundary conditions on $ u(x; \omega)$.
Suppose that $\hat{k}(x) \sim \mathcal{GP}(0, 4/25\exp(-(x-x')^2)) $.
Let $k(x; \omega)$ and $f(x;\omega)$ be two independent stochastic processes defined as
\begin{align}
	k(x) &= \exp [ \frac{1}{5} \sin ( \frac{3\pi}{2} (x+1) ) + \hat{k}(x) ],
	\\[5pt]
	f(x) &\sim \mathcal{GP}(\frac{1}{2}, \frac{9}{400} \exp(-25(x-x')^2)).
\end{align}

\begin{figure}[htb]
	\centering
	\includegraphics[height=4cm,width=16cm]{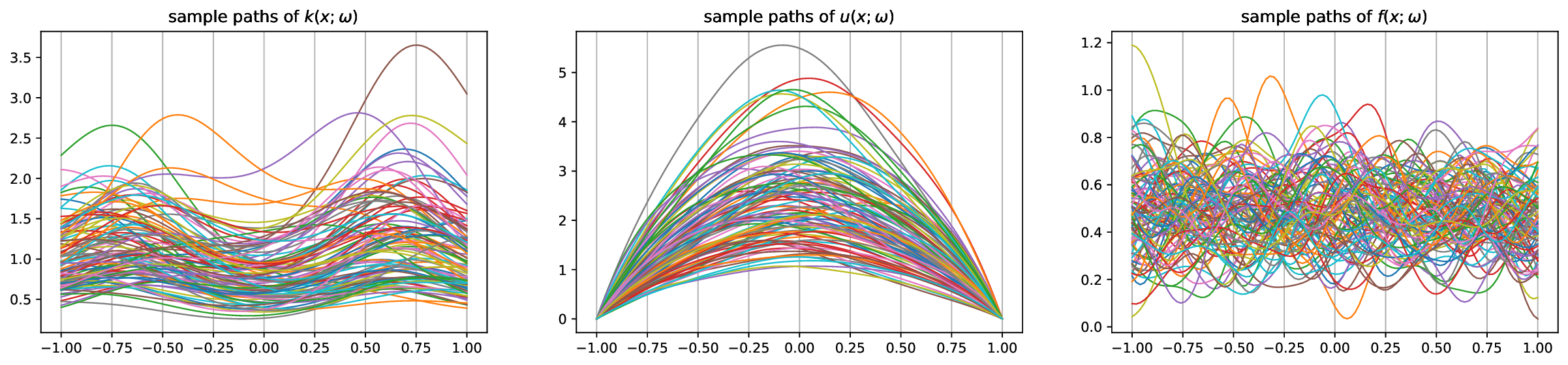}
	\caption{Partial training sample paths of $k(x; \omega)$, $u(x; \omega)$ and $f(x; \omega)$.}
	\label{sample_forward}
\end{figure}

We place 13 sensors and 21 sensors uniformly on $k(x;\omega)$ and $f(x; \omega)$,
respectively, and place 2 sensors on the boundary of $ u(x; \omega)$.
The training sample paths are illustrated \ref{sample_forward}.

To comprehensively analyze the effect of different settings, we conducted two sets of experiments.
In the first set, we maintain a constant total of 1000 training snapshots and  vary the dimensions of random noise as 2, 4, and 20.
Meanwhile, in the second set, we keep the dimensions of random noise fixed at 4 and alter the total number of training snapshots to 300, 1000, 2000, and 5000.

\begin{figure}[!htb]
	\centering
	\scalebox{.9}{\includegraphics[height=6cm,width=8cm]{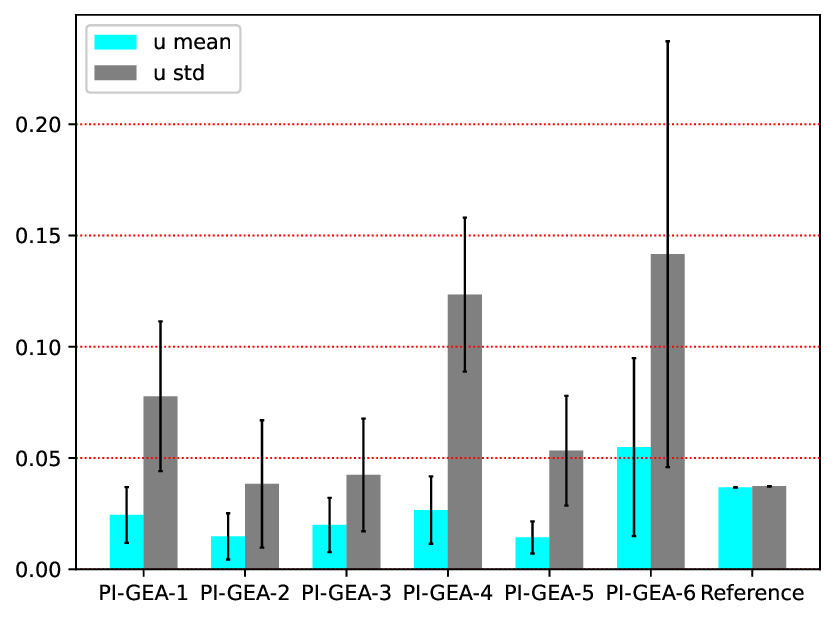}}
	\caption{Forward problem: relative $ L^2 $ errors under various settings.
		The bar chart and corresponding black lines depict the mean and standard deviation of the relative errors calculated from the 30 selected generators.
		(1) PI-GEA-1: 1000 training snapshots, with a batch size of 500 and a random noise dimension of 2.
		(2) PI-GEA-2: 1000 training snapshots, with a batch size of 500 and a random noise dimension of 4.
		(3) PI-GEA-3: 1000 training snapshots, with a batch size of 500 and a random noise dimension of 20.
		(4) PI-GEA-4: 300 training snapshots, with a batch size of 300 and a random noise dimension of 4.
		(5) PI-GEA-5: 2000 training snapshots, with a batch size of 1000 and a random noise dimension of 4.
		(6) PI-GEA-6: 5000 training snapshots, with a batch size of 1000 and a random noise dimension of 4.}
	\label{forward_error}
\end{figure}

Analyzing the insights derived from Figure \ref{forward_error}, we can draw several significant conclusions.
Firstly, while elevating the dimensionality of noise appears to enhance performance,
it is important to emphasize that achieving optimal outcomes is not assured.
Striking a balance between noise dimensionality and model performance becomes crucial to avoid overfitting or unnecessarily complex models.
Similarly, when the quantity of training snapshots is increased, a similar pattern emerges.
The need for more complex models to accurately capture the expanded data becomes evident.
This underscores the necessity of finding the right trade-off between the number of training snapshots and the model's complexity.
Such a balance is pivotal to achieving satisfactory and robust results.

\begin{figure}[htb]
	\centering
		\scalebox{.8}{\includegraphics[height=6cm,width=8cm]{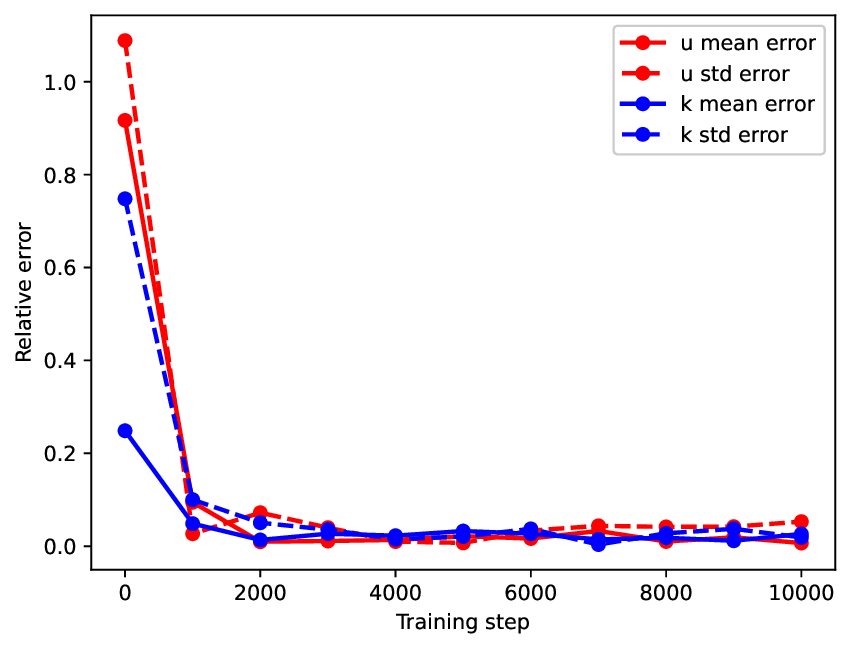}}
	\caption{Forward problem: relative error curves of PI-GEA during the training. Utilizing 1000 training snapshots, a batch size of 500, and random noise dimension of 4.}
	\label{forward_error_curve}
\end{figure}

Figure \ref{forward_error_curve} illustrates the relative error curve of PI-GEA during the training process.
For this particular visualization, we employed 1000 training snapshots, with a batch size of 500 and the random noise dimension of 4.
As the training advances, the relative error of the approximate solution consistently diminishes,
ultimately approaching zero.
This behavior signifies the model's continuous enhancement in terms of accuracy and performance over the course of training.

\begin{figure}[htb]
	\centering
	\scalebox{.9}{
		\includegraphics[height=6cm,width=14cm]{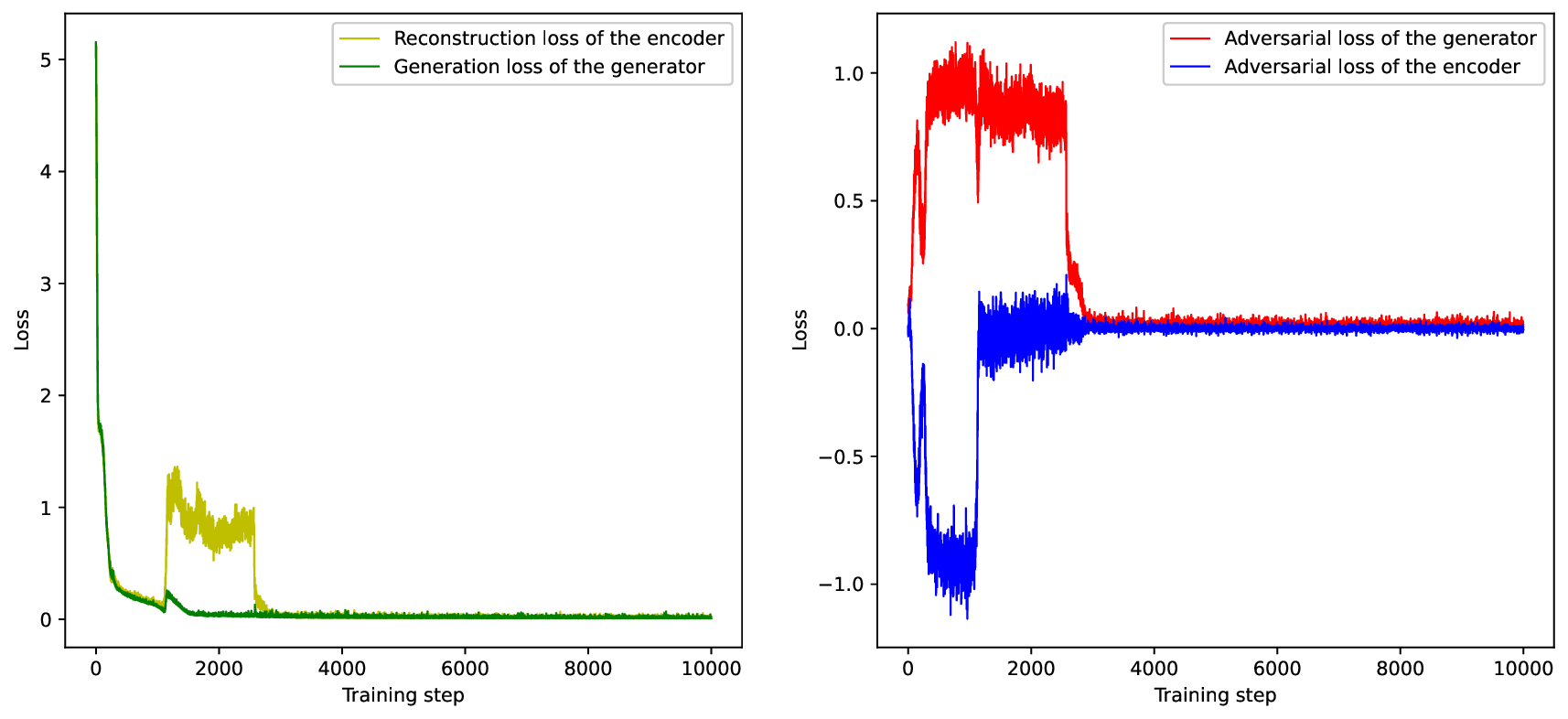}
			}	
	\caption{The training loss curve for each component of PI-GEA.}
	\label{loss}
\end{figure}

We further present the training loss curves for each component in Figure \ref{loss},
which serves as an example to verify the stability of our method.
As the training progresses, both the reconstruction loss and generation loss of PI-GEA approach zero,
indicating that the model is effectively reconstructing the data and generating accurate snapshots.
Furthermore, the adversarial loss of PI-GEA eventually reaches a balance after intense adversarial training,
demonstrating the stability of our method.
The convergence of the losses indicates that our approach is reliable and
capable of achieving stable and accurate solutions for stochastic differential equations.

\begin{figure}[htb]
	\centering
	\scalebox{1}{\includegraphics[height=6cm,width=8cm]{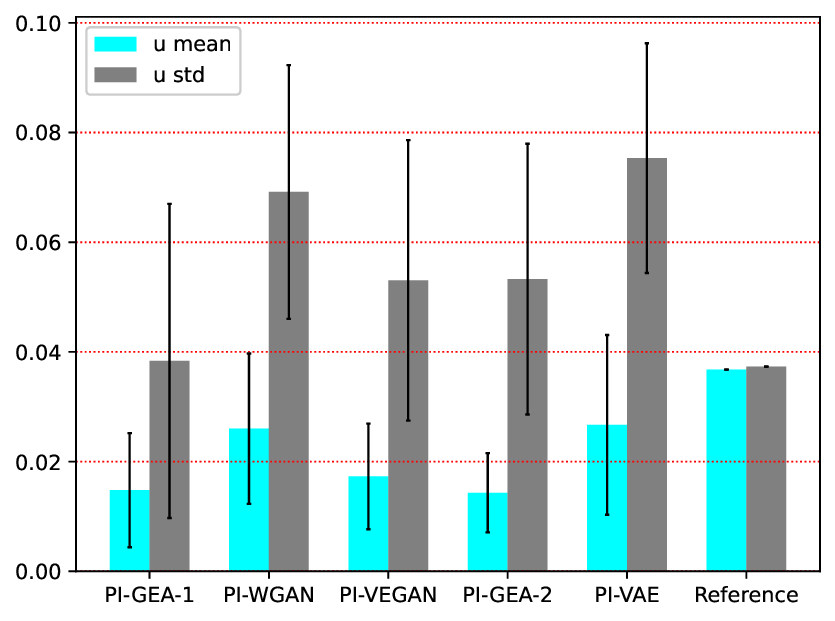}}
	\caption{Comparisons with the baseline methods for solving the forward problem.
    Here PI-GEA-1, PI-WGAN \cite{Liu2020}, and PI-VEGAN \cite{Gao2023Wang} utilize 1000 training snapshots,
	while PI-GEA-2 and PI-VAE \cite{Zhong2023} adopt 2000 training snapshots.}
	\label{forward_error_compare}
\end{figure}

At the end of this section,
we perform a comparative analysis of our approach against the baseline methods: PI-VAE \cite{Zhong2023}, PI-WGAN \cite{Liu2020}, and PI-VEGAN \cite{Gao2023Wang}.
The outcomes, presented in Figure \ref{forward_error_compare}, unveil compelling results.
Specifically, when the total number of training snapshots stands at 1000 and 2000, both PI-GEA-1 and PI-GEA-2 consistently outperform the compared baseline methods.
This comparison demonstrates the superior accuracy of PI-GEA in tackling the forward problem of stochastic differential equations.

\subsection{High-dimensional case}
\label{high_dim_problem}

In this section, we address a high-dimensional stochastic differential equation to assess the capability of PI-GEA
in tackling the issue of dimensional imbalance between the coefficients $k(x;\omega)$ and the forcing term $f(x;\omega)$.
Specifically,
\begin{align}
	\begin{split}
	k(x) &= \exp [ \frac{1}{5} \sin ( \frac{3\pi}{2} (x+1) ) + \hat{k}(x) ],
	\\
	\hat{k}(x) &\sim \mathcal{GP}(0, \frac{4}{25}\exp(-(x-x')^2)),
	\end{split}
	\\[5pt]
	f(x) &\sim \mathcal{GP}(\frac{1}{2}, \frac{9}{400} \exp(\frac{-(x-x')^2}{a^2})),
\end{align}
where $a$ denotes the kernel length scale of the stochastic process.
The default setting for $a$ is 1.
But when the correlation length of the forcing term $f(x)$ is relatively small,
the equation is referred to as a high-dimensional problem \cite{Liu2020},
which allows to assess the model's capability to handle the dimensional mismatch between the forcing term $f(x)$ and the coefficient $k(x)$.

We uniformly place 13 sensors on $k(x;\omega)$ and 2 sensors on the boundary of $u(x;\omega)$, respectively.
We investigate two distinct scenarios: (1) $a=0.08$ and (2) $a=0.02$.
In the case of $a=0.08$, we uniformly position 21 sensors across $f(x;\omega)$, while employing a random noise dimension of 10.
In the case of $a=0.02$, we uniformly position 41 sensors across $f(x;\omega)$, while utilizing a random noise dimension of 20.
For both scenarios, the total number of training snapshots is set to 5000, with a batch size of 1000.

\begin{figure}[!htb]
	\centering
		\scalebox{.8}{\includegraphics[height=10cm,width=12cm]{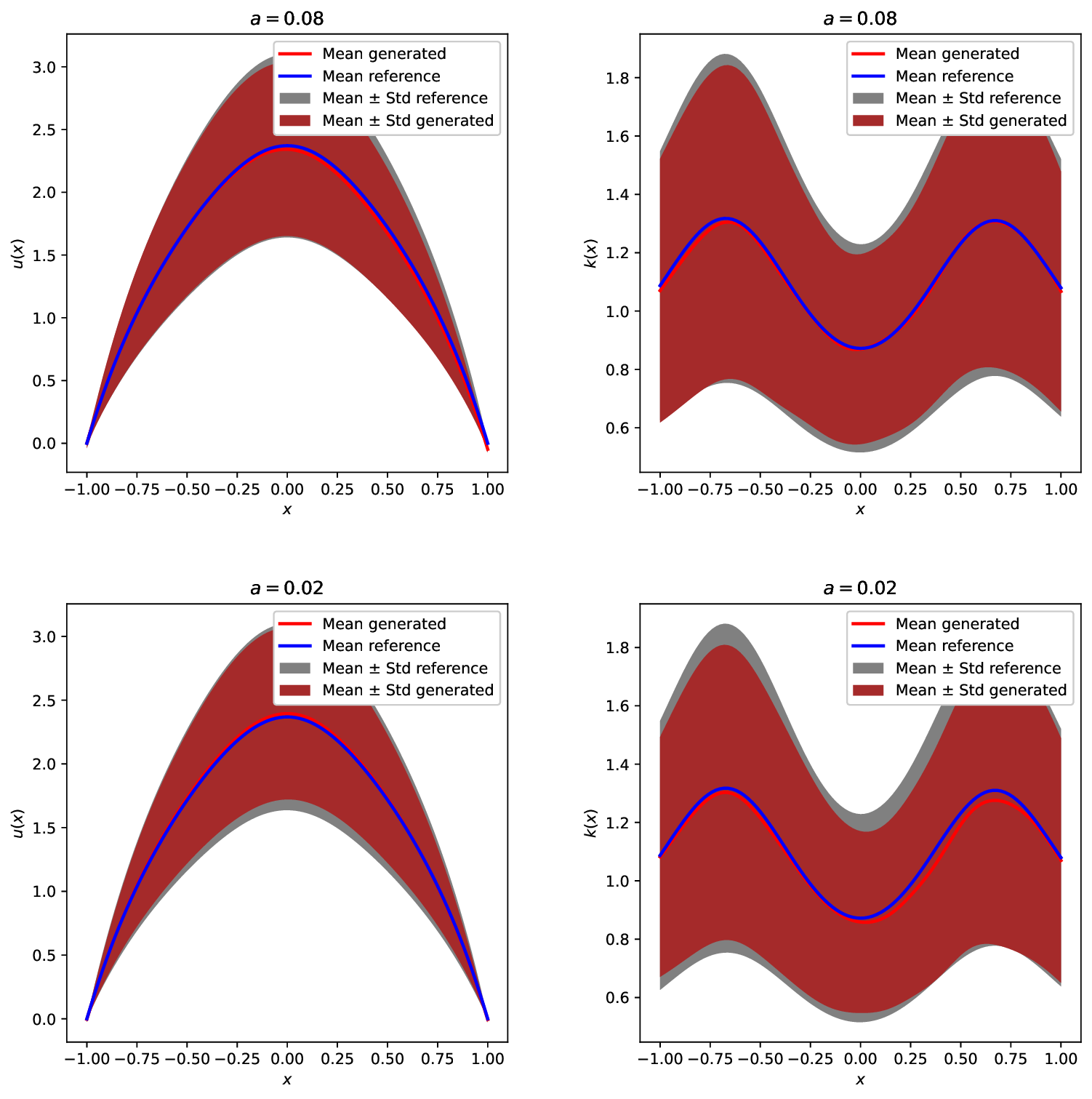}}
	\caption{Mean and standard deviation estimate of $u(x; \omega)$ and $k(x; \omega)$ using trained generator of PI-GEA for high-dimensional problems.}
	\label{high_dim_distribution}
\end{figure}

Our initial focus is to demonstrate the capability of PI-GEA in accurately reproducing the real distribution of snapshots.
Using the trained generator, we generate 1000 snapshots and compare them against 1000 real snapshots.
Figure \ref{high_dim_distribution} provides clear evidence that the distribution generated by our model aligns effectively with the true distribution,
showcasing commendable accuracy in both mean and standard deviation.

\begin{figure}[!htb]
	\centering
	\scalebox{.8}{
		\subfigure[$a=0.08$]{
			\includegraphics[height=6cm,width=8cm]{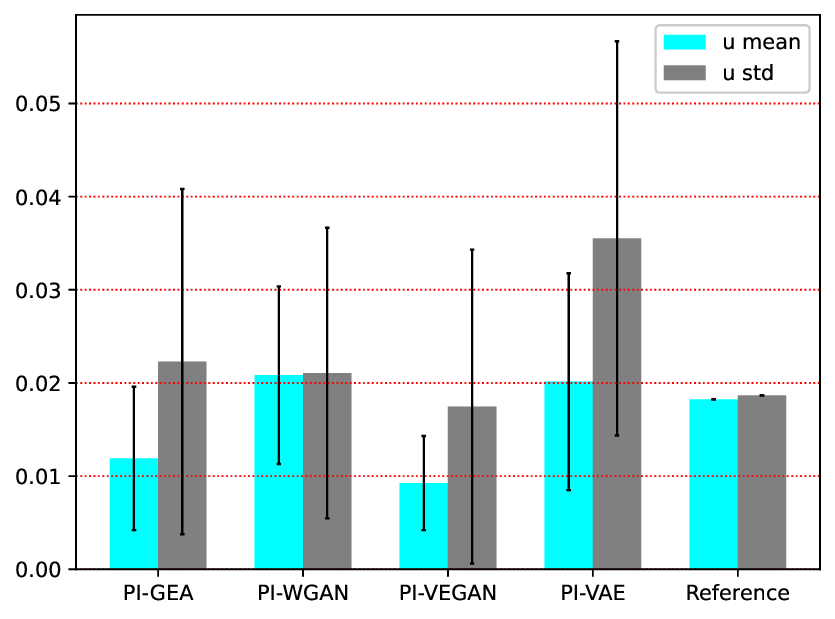}}
		\subfigure[$a=0.02$]{
			\includegraphics[height=6cm,width=8cm]{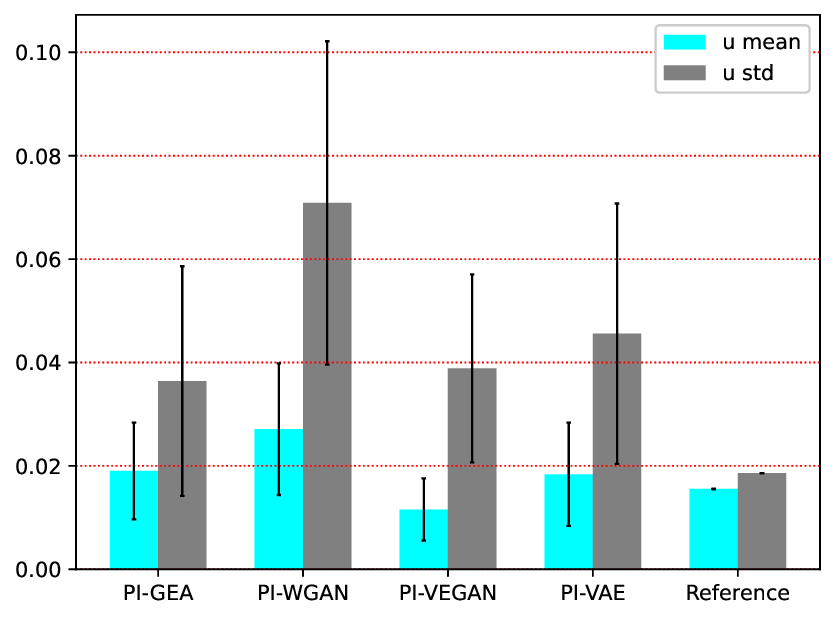}}
		}
	\caption{Comparisons with the baseline methods. Solving the high-dimensional stochastic partial equation using PI-VAE \cite{Zhong2023}, PI-WGAN \cite{Liu2020}, and PI-VEGAN \cite{Gao2023Wang}.
    For both scenarios, the total number of training snapshots is set to 5000, with a batch size of 1000.}
\label{high_dim_error}
\end{figure}

We further compare PI-GEA against PI-VAE \cite{Zhong2023}, PI-WGAN \cite{Liu2020}, and PI-VEGAN \cite{Gao2023Wang}.
As shown in Figure \ref{high_dim_error}, both PI-GEA and PI-VEGAN exhibit superior performance in terms of the relative $L^2$ error here.
This outcome signifies their high accuracy when tackling high-dimensional problems.
However, it is worth mentioning that PI-VEGAN entails greater complexity and demands a heavier computation cost.

\subsection{Inverse problem}
In this section we consider the inverse problem of stochastic differential equation \eqref{SDE_example}.
We uniformly place 1 sensor on $k(x;\omega)$, 13 sensors on $u(x;\omega)$ (including 2 sensors on the boundary), and 21 sensors on $f(x;\omega)$.

\begin{figure}[!htb]
	\centering
	\scalebox{.8}{\includegraphics[height=7cm,width=16cm]{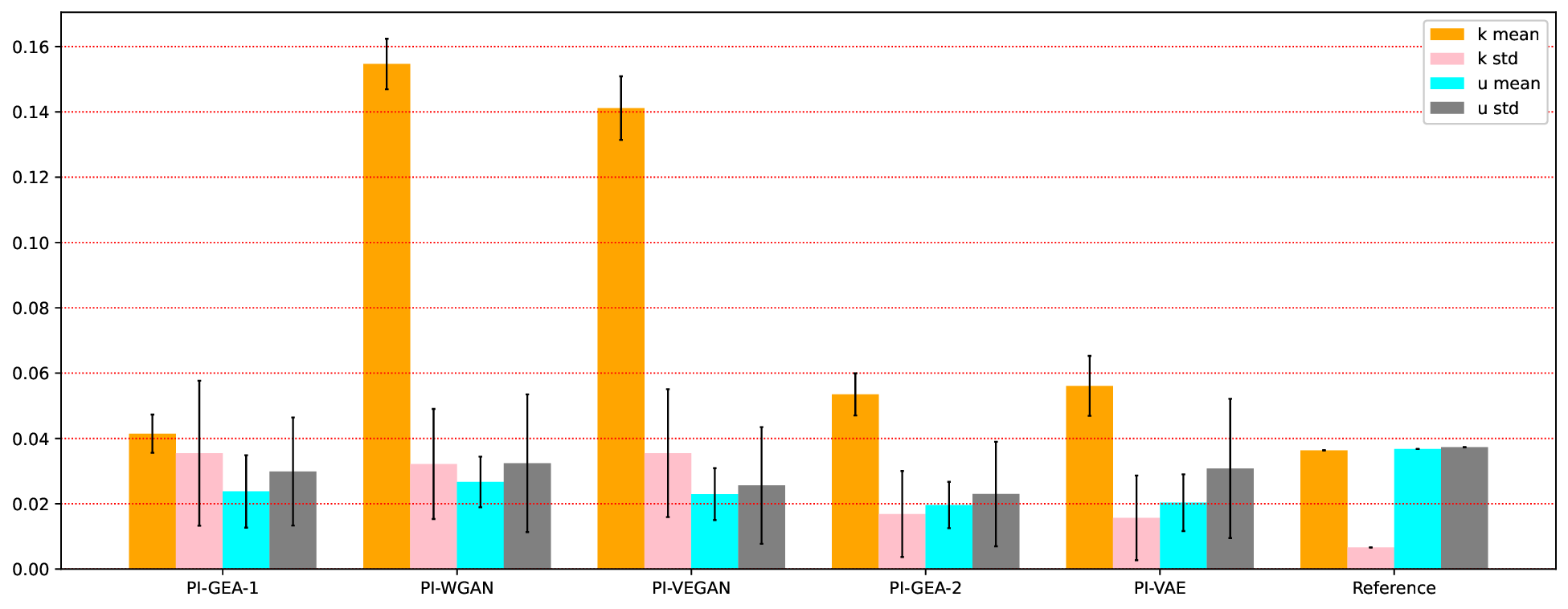}}
	\caption{Comparisons with the baseline methods for solving the inverse problem.
    Here PI-GEA-1, PI-WGAN \cite{Liu2020}, and PI-VEGAN \cite{Gao2023Wang} utilize 1000 training snapshots,
	while PI-GEA-2 and PI-VAE \cite{Zhong2023} adopt 2000 training snapshots.}
	\label{inverse_error}
\end{figure}

We compare PI-GEA with other baseline methods, and the results are presented in Figure \ref{inverse_error}.
From the observations in Figure \ref{inverse_error}, it can be seen that both PI-GEA and PI-VAE achieve superior performance in this problem.
With 2000 training snapshots, our model achieves a comparable or slightly better $L^2$ error than PI-VAE.
When the number of training snapshots is 1000, our model demonstrates a smaller relative $L^2$ error compared to PI-WGAN and PI-VEGAN, particularly for the mean of $k(x; \omega)$.
These findings underscore the effectiveness of our model in solving inverse problems.

\subsection{Mixed problem}

In this section, we evaluate the performance of the proposed PI-GEA in solving the mixed problem of the stochastic differential equation \eqref{SDE_example}.
We maintain 21 sensors for $f(x; \omega)$ and investigate two distinct scenarios:
(1) 15 sensors for $k(x; \omega)$ and 9 sensors for $u(x; \omega)$ (including 2 on the boundary).
(2) 9 sensors for $k(x; \omega)$ and 15 sensors for $u(x; \omega)$ (including 2 on the boundary).

\begin{figure}[!htb]
	\centering
	\scalebox{.8}{
		\subfigure[ 15 sensors for $k(x; \omega)$ and 9 sensors for $u(x; \omega)$.]{
			\includegraphics[height=7cm,width=16cm]{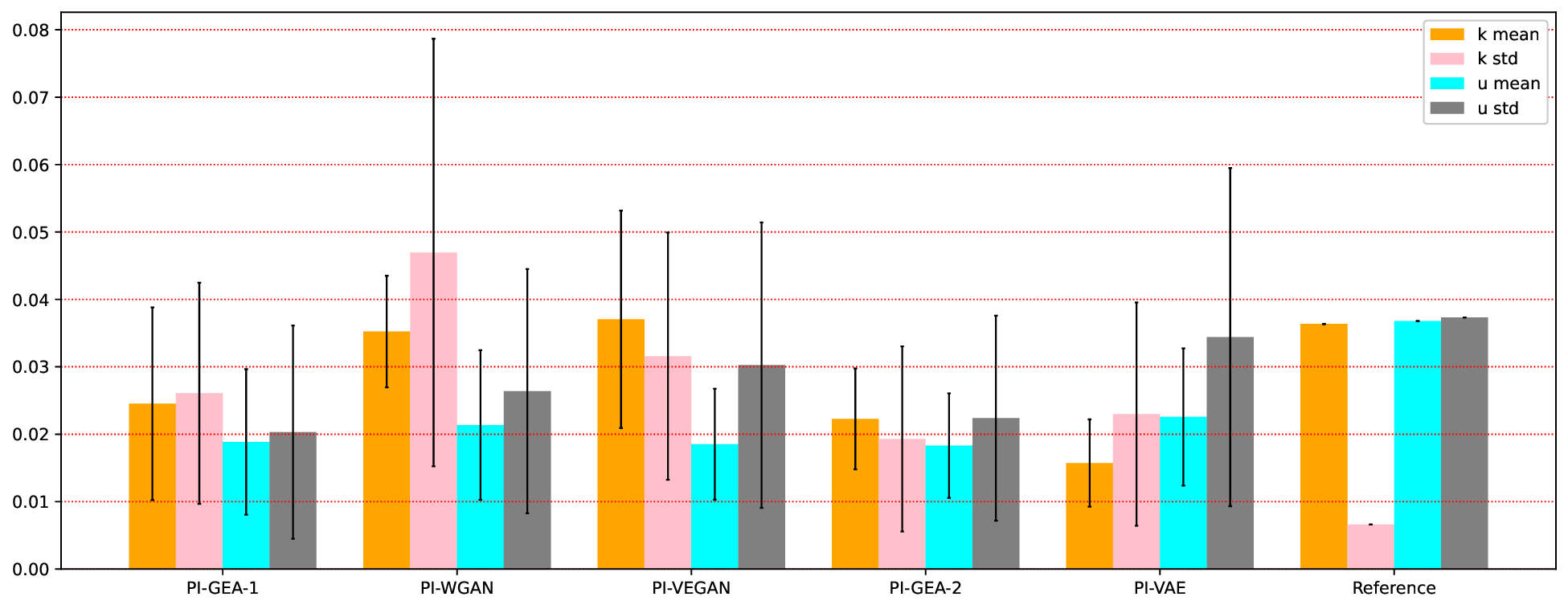}
	}}
	
	\scalebox{.8}{
		\subfigure[9 sensors for $k(x; \omega)$ and 15 sensors for $u(x; \omega)$.]{
			\includegraphics[height=7cm,width=16cm]{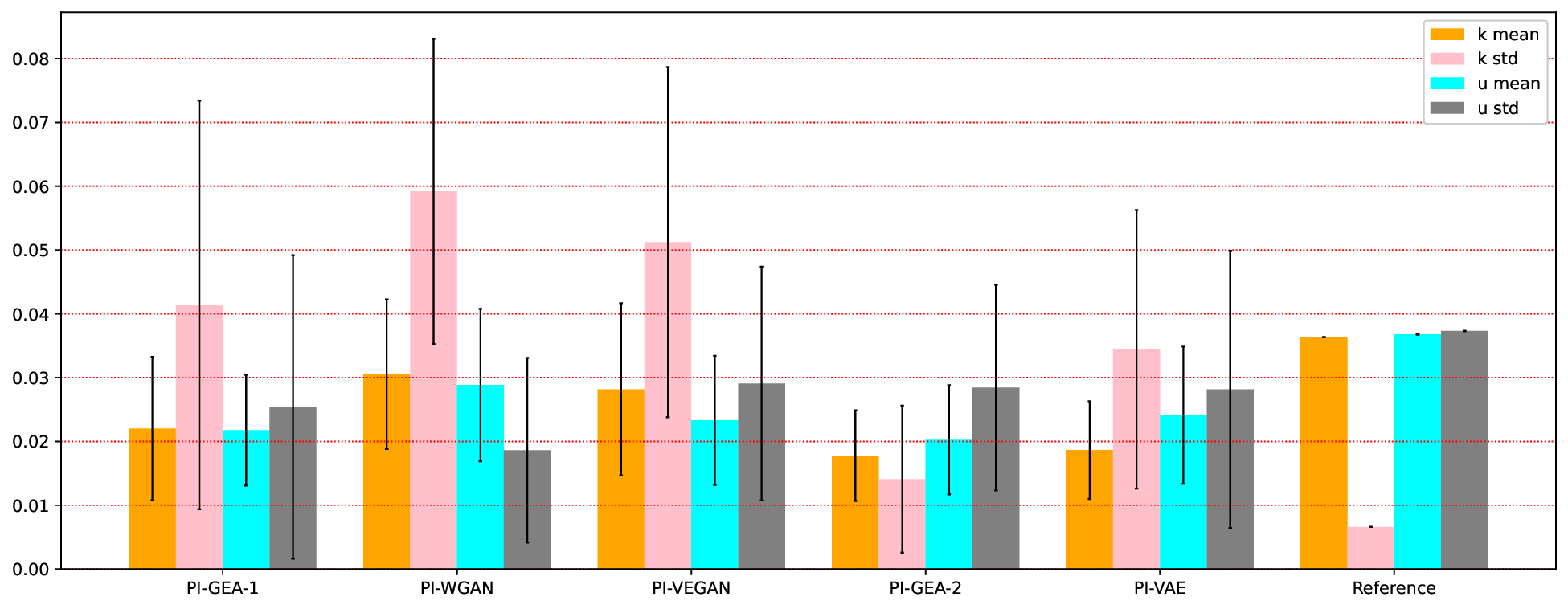}
	}}
	\caption{Comparisons with the baseline methods. Solving the mixed problem using PI-VAE \cite{Zhong2023}, PI-WGAN \cite{Liu2020}, and PI-VEGAN \cite{Gao2023Wang}.
    Here PI-GEA-1, PI-WGAN \cite{Liu2020}, and PI-VEGAN \cite{Gao2023Wang} utilize 1000 training snapshots,
	while PI-GEA-2 and PI-VAE \cite{Zhong2023} adopt 2000 training snapshots.}
	\label{mix_error}
\end{figure}

A comparison of the relative $L^2$ error of our model with other methods for solving the mixed problem is presented in Figure \ref{mix_error}.
In both cases (1) and (2), our model consistently attains the lowest relative $L^2$ error under identical conditions.
This outcome highlights the model's exceptional accuracy in addressing mixed problems.

\subsection{Computational Demand}
In this section, we conduct a comparison of model complexity and training time with PI-VAE \cite{Zhong2023}, PI-WGAN \cite{Liu2020}, and PI-VEGAN \cite{Gao2023Wang}.
We take the forward problem with $ a=0.08 $ from Section \ref{high_dim_problem} as an example.

\begin{table}[htb]
	\caption{Comparison of training time, model parameter count, and computational load across four models.}
	\label{model_param}
	\centering
		\begin{tabular}{|c|c|c|c|}
			\hline
			& Time per epoch/s & Params/M & Flops/KFlops \\	
			\hline
			PI-VAE \cite{Zhong2023}
			& 0.886 & 0.06 & 59.39 \\
			\hline
			PI-WGAN \cite{Liu2020}
			& 3.772 & 0.09 & 89.73 \\
			\hline
			PI-VEGAN \cite{Gao2023Wang}
			& 4.256 & 0.11 & 113.28 \\
			\hline
			PI-GEA
			& 3.470 & 0.06 & 59.39 \\
			\hline
	\end{tabular}
\end{table}

The insights presented in Table \ref{model_param} are highly informative.
It is important to emphasize that PI-VAE relies on the variational auto-encoder structure and does not involve adversarial training,
thus resulting in its better performance in terms of training time.
However, our method excels in accuracy as shown in previous sections.
Conversely, PI-WGAN \cite{Liu2020}, PI-VEGAN \cite{Gao2023Wang}, and our approach all adopt adversarial training frameworks.
Nevertheless, when compared to PI-WGAN and PI-VEGAN,
our model showcases advantages across all three dimensions: training time, model complexity, and computational efficiency.

\section{Conclusion}
The paper acknowledges the limitations of current deep learning methods for sloving SDEs,
which primarily focus on minimizing the difference between approximated solutions and real snapshots.
However, this intuitive approach faces challenges due to the complex nature of data distribution.
Inaccuracies arise from this complexity, and training instability emerges from the interplay between high dimensionality and intricate data distributions.

We introduce a new class of physics-informed neural networks that employs an alternative strategy.
This strategy involves matching in a low-dimensional latent space and incorporates distributional analysis to guide the optimization process.
The proposed approach deviates from the conventional direct fitting strategy, addressing challenges related to data complexity and training instability.

Our approach demonstrates significant accuracy advantages compared to existing deep learning solvers for three types of SDE-related problems:
forward, inverse, and mixed problems where both system parameters and solutions are simultaneously calculated.
This result underscores the practical relevance and effectiveness of the presented approach in various scenarios.

The introduced model also exhibits superiority over existing adversarial deep learning solvers along three dimensions: training time, model complexity, and computational efficiency.
This highlights the advancements achieved not only in accuracy but also in practical usability and efficiency.

It is worth noting that the proposed model assumes noise-free sensor data,
which might not hold true in real-world scenarios.
To address this limitation, future research can focus on developing noise-robust models that
can effectively handle stochastic differential equations in the presence of noise and perturbation.
Such advancements will enhance the applicability and reliability of the approach in practical applications.



\end{document}